# Structured Differential Learning for Automatic Threshold Setting

Jonathan H. Connell, Benjamin W. Herta
IBM T.J. Watson Research
19 Skyline Dr., Hawthorne NY, USA
`{jconnell, bherta}@us.ibm.com`

## Abstract

*We introduce a technique that can automatically tune the parameters of a rule-based computer vision system comprised of thresholds, combinational logic, and time constants. This lets us retain the flexibility and perspicacity of a conventionally structured system while allowing us to perform approximate gradient descent using labeled data. While this is only a heuristic procedure, as far as we are aware there is no other efficient technique for tuning such systems. We describe the components of the system and the associated supervised learning mechanism. We also demonstrate the utility of the algorithm by comparing its performance versus hand tuning for an automotive headlight controller. Despite having over 100 parameters, the method is able to profitably adjust the system values given just the desired output for a number of videos.*

## 1. Introduction

Machine learning techniques have gained popularity in the computer vision field recently. By using large volumes of data the hard issues of feature discovery and combination can often be finessed. However the price that is usually paid is to have a "black box" controller at the center of the system. By this we mean something like an SVM or neural network whose internals are hard to interpret intuitively. Moreover there is no good way to manually "tweak" systems built in this fashion. Structured Differential Learning (SDL) is a tool that overcomes this by working with conventional rule-based vision system designs. That is, the system can be composed of any number of discrete parts, each with its own recognizable teleology. Generally the SDL learning technique to be described will work with **any** system that is a collection of thresholds, combination logic, and time constants.

When there are parameters with clear meanings (to humans) it is possible to initially bias the system into an approximately correct configuration. It is also possible to alter parameters by hand in a reasoned way to slightly alter the system's behavior. Unfortunately the proliferation of free parameters in such systems make them quite difficult to fully optimize by hand. While Monte Carlo search, genetic algorithms [2], stochastic approximation [3], gradient estimation [4], or expert systems [5] can be used to explore small regions of parameter space, such approaches are expensive either computationally or data-wise. For instance, in gradient estimation the system must be run over the whole dataset a large number of times, each iteration having small changes in some selected parameters. The SDL method, by contrast, is very quick but must be given a starting point which is "close" to the ideal solution. It cannot simply be loaded with random values and then be expected to converge on the global maximum.

In operation, SDL attempts to attribute each output error to a single cause. While there may be many possible adjustments or sets of adjustments that will correct an error, spreading the examples among many causes (as in back-propagation [6]) defocuses the system and makes it slower to converge. Also it is assumed that the system already performs reasonably well, so adjusting a large number of parameters simultaneously is probably a mistake. In addition to the single cause hypothesis, SDL also uses the "near miss" [7] heuristic: if a proposed change is too large it should be rejected. Instead the system will wait until it receives a less ambiguous example. Generally this combination of restrictions lets SDL arrive at a good set of parameters with less labeled data than other methods.

## 2. Forward system

There are two main parts to the solution. The first is a library of control system pieces which keep track of possible near misses/hits in the parameter space. The second piece is a method of using the near miss/hit data in combination with the system result and ground truth data to determine how to alter the control parameters.

Consider the example in Figure 1 which shows how a detector for far away taillights in a headlight control system might be implemented. Given a set of bright spots in the input image, the system first checks properties of each spot found against specified passbands. For instance, it might check that the spot is near the horizon in the image, that it is greater than 4 pixels but less than 30, etc. Then, if the spot passes all the tests, it is considered a plausible candidate to be the far taillight of some vehicle. If this output is wrong – the system either missed some taillight or found an extraneous one – the system must figure out which parameter(s) to adjust, and in which direction.

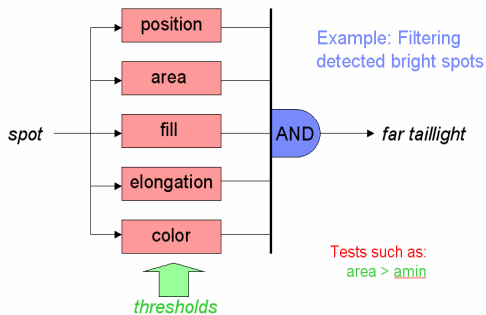

Figure 1: Conventional structured vision systems often work by combining feature tests with logical primitives.

## 2.1. Soft thresholds and combination logic

The most basic processing element is a threshold unit. Conceptually this returns "true" if a value is at or above some threshold value, and "false" otherwise. A similar threshold unit can be built that returns "true" when the value is below some limit, instead. Such decisions can be "softened" so that they instead return a "score" based on how close the value is to the threshold. Generally this only occurs over a particular range, with saturation beyond these "tolerance" limits. That is, the comparison is clearly true or false – it is not reasonable to adjust the parameter beyond these points. This leads to a scoring function with a single slope around the current setting, similar to a fuzzy logic predicate [8]. We can also directly determine an "alternate" shifted value for the threshold parameter that will cause the current input to either pass or fail.

Graded response values can be combined in a number of ways. The simplest are logical AND and OR. These can be implemented (again, as in fuzzy logic) using a minimum and maximum function, respectively. Two threshold modules can then be used together on a single value to simulate "bandpass" filters (with AND) and "notch" type filters (with OR). Moreover, multiple criteria, such as constraints on a variety of different measurements, can be combined using AND as shown in Figure 1. Similarly, multiple different independent justifications for the same result can be combined using OR.

A big difference from fuzzy logic is that these combination methods also attempt to record the most "defeasible" of their inputs (cf. belief revision in a truth maintenance system [9]). The AND module records which input had the lowest value. If the result was true, then changing just this "suspect" input to false (by using the associated alternate value) will change the overall decision to false. However, if two or more inputs are false then there is no clear choice (as in test vector generation [10]) and the decision is indefeasible. That is, we cannot do credit assignment backwards through the AND in this case. Or it may be that while there is a clear input to change, its score is saturated (i.e. unalterably false). Again, there is no reasonable single parameter to adjust to change the overall decision. However, usually this is not a problem in practice because we have thousands of other labeled data points (e.g. >200K images in Section 5) to make the proper action clearer.

For OR we do a similar thing but record the strongest input. If there is only one true input, then changing this to false will change the overall result to false also. However, if there two or more true inputs, or a saturated true value, then the function is no longer differentiable: there is no single pivot point ("suspect" parameter) for the decision.

With graded thresholds, AND, and OR we can directly implement the piece of control logic shown in Figure 1. Moreover, in the case of an incorrect decision we can usually identify a single most readily changeable parameter to adjust by propagating the error back through the logic.

## 2.2. Time constants, counting, and regions

We can combine not only values from different subsystems, but also from the same subsystem over time. Figure 2 shows a sequence of decisions over a short time interval. By feeding these to an AND gate using a sliding window we can make sure that some condition is true for at least V consecutive time frames (5 in the diagram). As before, the AND gate will derive its reported score, suspect, and alternate value from the weakest of these time instants.

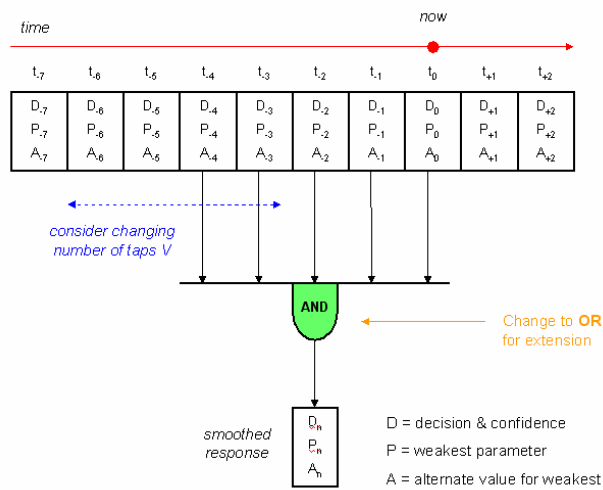

Figure 2: The temporal filtering function can be viewed as an AND over successive time instants to reject spurious detections.

However, we can also contemplate what would happen if we asked instead for only 4 consecutive occurrences, or perhaps 7 instead. To determine a useful alternate value we scan the inputs for a small range past the nominal time constant to find the first false input. That is, for the example above where V=5, we scan from $t_0$ to $t_{-7}$. Suppose the first false was found at $t_{-4}$, then the output of the AND gate as shown would be false. However we could make it be true if we instead only considered the first V=4 elements ($t_0$ to $t_{-3}$). Now suppose that the first false input is found at $t_{-6}$ instead, which means the output of the AND gate is true ($t_0$ through $t_{-4}$ are all true). If we want to change this to false we can set V=7 to include this first false input.

In both these cases we can then use a ramped threshold centered at V=5 to assign a confidence to such a change. It may be that altering V is less "expensive" than adjusting any of the inputs. In such a case the parameter V itself would be reported as the suspect in the output along with the appropriate alternative number of taps for the AND gate. This is a particularly useful fall-back position if none of the inputs is defeasible (i.e. if there are many false or one stuck at true with very high confidence).

Another useful temporal construct is the retriggerable "monostable" which serves to extend an output over some length of time even if the triggering signal is not currently present. In this way the system can ignore short data dropouts. In practice we often employ this extension operation in combination with temporal smoothing on the input. Since the output stays true for so long, we want to make sure it is not triggered by spurious noise.

The implementation of a monostable is very similar to the temporal smoothing element described earlier, except it uses OR to combine inputs over time. As with the temporal smoother, alternative values for the time constant are also considered and assigned confidence values. So, if a change of output truth value is desired, the monostable may suggest changing its own time constant rather than altering any parameter associated with its inputs. There is, however, one difference that has proved to be useful in practice. In the case where multiple inputs are true, a normal OR gate would not be able to find any one input that could be changed to change the overall result. We relax this for the monostable and instead rotate blame among the various true inputs. Otherwise it is very hard to propagate error backwards through a monostable – the time constant ends up being blamed in most instances instead.

Not only can outputs from a single module be combined over time, but outputs from the same module operating on different inputs can be combined. For instance, in a headlight control system we might have a module that determined whether some bright spot was a streetlight. If we declare an urban area to be anywhere which has 3 or more streetlights, we need to combine the output of the streetlight detector running on each of the bright spots found in the image. We can do this by building a list of objects ranked by our confidence that they are streetlights, and then seeing if the first N of them pass (are true). This is essentially connecting an N input AND gate to the first N elements in the list. The score of the third weakest candidate (if any) then gives the score for the "at least 3" counting block.

As with the temporal smoother we can also explore how the output changes if we alter the number of items required. For instance, in our example it may be more desirable to change the number of streetlights from 3 to 2 rather than alter any aspect of the streetlight detector itself. There are several ways the raw count information could be combined with the threshold adjustment information. Our preferred method is to take as the score the most defeasible of the two scores. Then, since threshold adjustment is a meta-level operation, we keep the original "suspect" parameter if one of the candidates was defeasible. If none of the spots in the original count were defeasible, we instead record the threshold itself as the "suspect" parameter. This combination method lets the "blame" percolate backwards through the counting process rather than pinpointing the count threshold itself most of the time.

Comparison modules are not limited to scalars but can also be built to works on value histograms. A histogram module might generate a graded response based on the fraction of data below a certain value threshold. For instance, the module might test whether or not 30% of the pixels in a region of the image were below a gray value of 50. The raw fraction could then be passed through a single-ramp threshold function running, say, from fully false at 25% to fully true at 35%. At the meta-level the threshold itself could also be assessed as a potential "suspect" parameter. That is, what if the threshold on the pixel count was set to 20% or 40% instead?

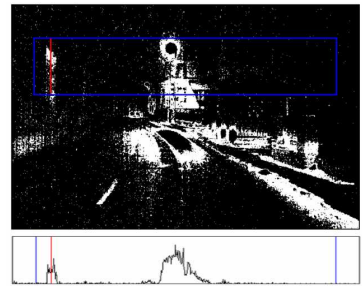

Figure 3: The limits of a region can be tuned by looking at the projection of the thresholded pixels then moving the edge of the region to adjust the total count so it is now either under or over.

Modules can also be built for spatial projections. This is useful for evaluating rectangular regions of images as shown in Figure 3. Suppose the histogram in the previous example was computed over just a particular rectangular region of the image with, say, 100 < x < 200 and 50 < y < 100. What would happen if the lower x limit was changed to 95 instead? This meta-level adjustment can be evaluated in the same way as the histogram threshold discussed above, thus letting the x limit of the rectangular region become a potential "suspect" parameter also.

## 3. Learning procedure

During operation of the system a network of various modules as described works together to produce an output decision along with bookkeeping information about the most "suspect" parameter involved in each decision. As part of the bookkeeping we also record a proposed alternative value for the parameter to potentially change the overall decision. This becomes vital information for the second half of the scheme – the method for actually adjusting the parameters to obtain better performance.

### 3.1. Histogramming

If we have groundtruth for each binary decision the system responses can be assigned to one of 4 possible categories: true positives (TP) where both said "true", true negatives (TNs) where both said "false", false positives (FP) where the system erroneously said "true", and false negatives (FNs) where the system erroneously said "false". Typically, some of the results are defeasible while others are not. We are only concerned here with the defeasible ones – particularly their designation of a suspect parameter and the associated alternative value. In Figure 4 we histogram the 4 categories of decisions separately for each parameter, where the x axis is the alternative value and the y axis is how many instances proposed this value. The current value is delimited by brackets (and obviously has no instances).

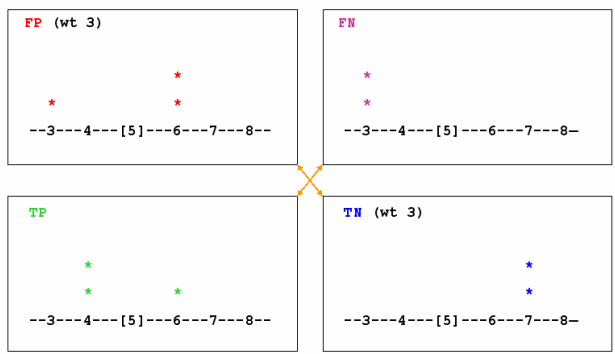

Figure 4: The first step in learning is to classify each result and histogram them based on the alternate value for each parameter..

Generally, errors of various types are weighted differently: FPs might be 3 times as damaging as FNs. We can take this into account by multiplying both the FP and TN histograms by 3. We then look at the effect of changing the parameter to one of the alternative values. In the case of a FP event, the change will transform this into a TN. Conversely, if it was a TP event, it will be transformed to a FN event. We capture this by creating a single combined histogram by subtracting off the correct events from the wrong events. Thus the amount of correction or "benefit" provided by each alternative parameter value is explicitly tabulated. Figure 5 (left) shows this for the values from Figure 4 where B = (FP + 3 * FN) – (TP + 3 * TN).

The actual effect of changing the parameter to some other value is the sum of all benefits between the original value and the new value. To help catalog this we generate a cumulative histogram of the benefit function away from the original value as shown in Figure 5 (right). The highest value of this graph is thus the value of the parameter that will likely produce the best reduction in error rate. If there are several values that are tied, then the value closest to the original setting is selected. In Figure 5 the original value was 5 (bracketed), but it can be seen that 6 is the best choice for highest system performance. This value fixes the most errors while introducing the fewest new ones (with weighting taken into account).

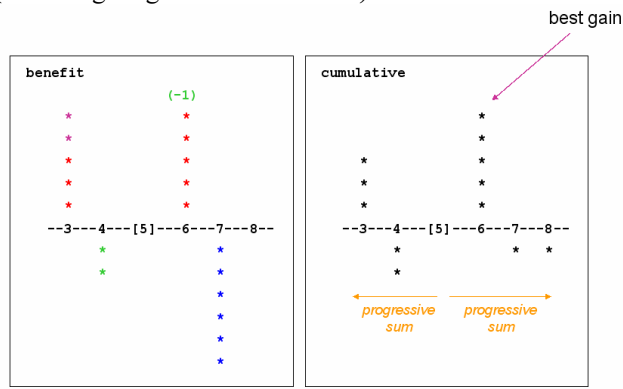

Figure 5: The histograms are combined bin-wise and then integrated outwards from the current value to find the best gain.

Unlike this simple example, real benefit curve are not so sharp and often gradually asymptote. To prevent unnecessarily extreme changes from being made, we usually pick the closest bin to the original value that yields a cumulative benefit of at least 90% of the maximum.

### 3.2. Updating parameters and tolerances

Technically, the predicted benefit for changing some parameter is only valid if it is the *only* parameter altered. Yet we typically assume that the parameters of the system are either independent or minimally coupled. This rationalization authorizes us to adjust multiple parameters on each learning round. Note that generally we perform several such rounds of optimization, reprocessing the original data with the new parameters to generate new outputs. In light of this, any unintentional interference between adjustments can be automatically mitigated in the next iteration.

While we could adjust *all* parameters each time, we usually update only a small fraction. In practice there are often very few counts in the histograms for most parameters. Making adjustments based on such sparse noisy data is usually a mistake. Thus we prefer to only adjust parameters where the new setting will fix at least a certain number of events (such as 10). The number of events fixed can be found by performing the summation and accumulation described above, but with FPs and FNs weighted equally.

The actual use of SDL is typically in a cooperative human/machine optimization setting. We first set, by hand, all the parameters to what we believe are reasonable values based on our past experience. We then run SDL for several iterations (less than 10) and pick the best operating point found. From there we manually examine the errors produced and try to make changes by hand to fix them. Generally we make the smallest manual changes possible so that we do not "break" the system's performance on other video situations we did not directly examine. These changes serve to "kick" the system into a better attractor basin for SDL. We then repeat the whole combined adjustment cycle as needed to arrive at a final configuration.

Note that the whole credit assignment process as described depends crucially on the tolerance ranges assigned by the human designer. A parameter with a large tolerance has the ability to "hog" all the credit (and blame) since it is "easy" to change its value. This may mask other parameters which are the real cause of the remaining problems. Or, for a system with multiple decision paths combined with OR, one *path* might "hog" all the training data. Conversely, a parameter with too small a tolerance will seldom be the "suspect" and hence have too few points in its histogram to be considered for alteration at all.

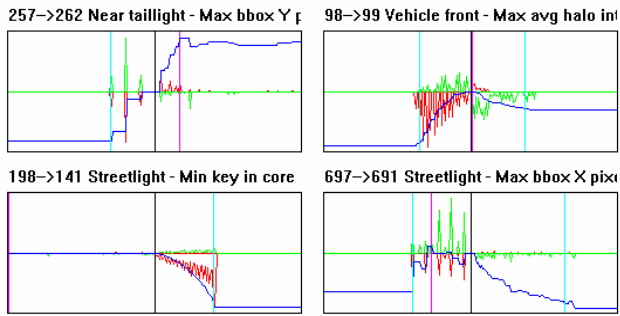

Figure 6: The tolerance of the soft threshold can be adjusted by looking for a substantial drop (cyan) from the best value.

Fortunately we can automatically estimate parameter tolerances and thus ameliorate these situations. Figure 6 shows how we use the same cumulative benefit curve (blue) to pick new tolerances. Here, markers (cyan) show where the benefit drops to 90% of the lowest value on either side of the peak (magenta). Conceptually, as the parameter is moved from the peak to this limit the benefit will go from its maximum to its minimum. This is exactly the intuitive idea behind the tolerance – describing the range over which it is reasonable to adjust the parameter. The positive and negative tolerances are adjusted separately to accommodate dual slope thresholds.

Empirically we change the tolerance to be 2.5 times this distance (or leave it unaltered if a suitable drop point is not found). The factor of 2.5 is to allow the tolerance to be increased even further in the next round, since the histograms can only be plotted over the tolerance range (there is no data generated elsewhere). Of course, as with the parameter value itself, if there is too little information in the histogram then no change is made to the tolerances.

## 4. Experiments

To provide some evidence of the value of SDL, we report on experiments performed with an Intelligent Headlight Control (IHC) system (see [1] for details). This system uses a forward facing video camera to automatically switch between bright and dim headlights. For training we collected several hours of video data and annotated what the correct headlight decision should be for each frame.

We used several metrics to assess performance of the end-to-end system. The most basic is how many times the headlight changes were too early, too late, or were completely unjustified (e.g. a "blip" to highbeams). In these cases the headlights are momentarily either too bright or too dark. We normalize the counts for both types of events against the ideal number of changes and express the result as a percentage (e.g. "too dark / switch"). We want these error percentages to be as low as possible. We also look at a combined statistic which is the percentage of correct transitions versus incorrect ones ("ok / wrong") which should be as high as possible. Finally, we count special "blinding" events where the car's highbeams were still on despite the presence of a very close oncoming vehicle. We want the blindings per hour to be as small as possible.

Note that SDL is just trying to get the correct highbeam decision on each video frame. It can only improve these other metrics indirectly – it cannot utilize these particular error values in its tuning procedure. Yet, despite this mismatch, it can still make significant improvements (partly due to the temporal smoothing and monostables used in the actual IHC system).

We report here on experiments from two different phases of the development of the IHC system. In an early phase we had a system with 171 adjustable parameters. This was hand-tuned to the best of our abilities to yield version Hand-1. We then ran SDL training with 1.9 hours of video (200K images) to yield version SDL-1. The second experiment comes from a later development phase in which the prototype system had 228 free parameters.

Again, Hand-2 was the best configuration we could arrive at with hand tuning. Using SDL training with 12.6 hours of annotated video (1.3M images) we built version SDL-2.

Table 1 shows the results for the hand trained systems and their SDL counterparts, both on the video training sets and in a live test drive. Testing the hand tuned system on the training videos is reasonably legitimate since only a small number of problem sequences were actually examined by the human. The results of the SDL system on its own training data are not a good predictor of actual performance, but they at least show how well it was able to fit the data set.

| version | too bright / switch | too dark / switch | ok / wrong | blind / hour |
|---|---|---|---|---|
| **Hand-1** | **47** | **30** | **66** | **3.8** |
| **SDL-1** | **30** | **33** | **82** | **2.7** |
| *Hand-1 in car* | *24* | *18* | *195* | *3.0* |
| *SDL-1 in car* | *16* | *24* | *183* | *0.0* |
| **Hand-2** | **30** | **28** | **58** | **5.5** |
| **SDL-2** | **18** | **36** | **69** | **2.1** |
| *Hand-2 in car* | *29* | *8* | *218* | *10.3* |
| *SDL-2 in car* | *8* | *20* | *313* | *1.0* |

Table 1: Performance of two hand tuned systems and their SDL updated counterparts. Results are shown for both the off-line video collections and for the in-car test drives.

The in-car tests for both type of systems are the least objectionable, but somewhat subjective. Here the IHC system was allowed to control the headlights while a human with a clipboard tallied early and late transitions, as well as spurious (de)activations and blinding events. While the car was driven in the same area where the training data was acquired, the details of traffic, weather, and road conditions were not identical. Each test drive was approximately one hour long (100K decisions) and tested only one system at a time (so the actual required responses were different).

As can be seen, the SDL versions outperform their starting manual versions in both situations based on a lower total number of error events (too-bright + too-dark) and the important blindings / hour statistic. Interestingly the trade-off selected by the human and SDL relative to the too-bright events versus too-dark events is different. Also note the dramatic decrease in the number of blindings of version SDL-2 relative to Hand-2. This is one case where it certainly pays to have an automated procedure go through *all* the training data.

Given the fixed structure of the IHC system, without SDL we would be stuck with the performance that could be obtained by hand tuning. While SDL is in some sense ad hoc and not guaranteed to converge, it at least gives us an automatic "button" we can press to improve the results.

## 5. Conclusion

We have shown how to soften threshold decisions using tolerances, and how to use the resulting scores to identify the weakest input to combinational logic functions. We went on to show similar credit-assignment schemes for counting, time constants, and region specifications. With this framework we can compare the final Boolean result of a controller to a desired state, and then percolate the credit or blame for the decision all the way down to a single parameter. By histogramming errors against alternate parameter values for a large number of decisions we can pick better values for the parameters, and also re-estimate the threshold tolerances. We validated the utility of this method in the context of an automotive headlight control system and demonstrated significant gains over the best human tuning.

All in all, SDL provides a hitherto unavailable method for efficiently tuning conventionally structured computer vision system (even potentially legacy ones). We are currently investigating the application of SDL to other problem domains, and working to extend the technique to regression, vectors, and set-based values.